\pdfoutput=1
\documentclass{article}
\usepackage{iclr2021_conference,times,jamie}
\usepackage{tikz}
\usetikzlibrary{positioning}
\usepackage{graphicx}
\usepackage{listings}
\let\origurl\url
\renewcommand{\url}[1]{\penalty10000 \hskip.5em
    plus\linewidth \interlinepenalty10000\penalty200
    \hskip-.17em plus-\linewidth minus.11em \origurl{#1}}

\title{
  Lossless compression with state space models using bits back coding
}

\author{James Townsend\\
Department of Computer Science\\
University College London\\
\texttt{james.townsend@cs.ucl.ac.uk}\\
\And
Iain Murray\\
School of Informatics\\
University of Edinburgh\\
\texttt{i.murray@ed.ac.uk}
}

\iclrfinalcopy
\begin{document}

\maketitle

\begin{abstract}
  We generalize the `bits back with ANS' method to time-series models with a
  latent Markov structure. This family of models includes hidden Markov models
  (HMMs), linear Gaussian state space models (LGSSMs) and many more. We provide
  experimental evidence that our method is effective for small scale models,
  and discuss its applicability to larger scale settings such as video
  compression.
\end{abstract}

\section{Introduction}
Recent work by \citet{townsend2019} shows the existence of a practical method,
called `bits back with ANS' (BB-ANS), for doing lossless compression with a
latent variable model, at rates close to the negative variational free energy
of the model (this quantity bounds the model's marginal log-likelihood and is
often referred to as the `evidence lower bound', or ELBO)\@. BB-ANS depends on
a last-in-first-out (LIFO) source coding algorithm called \emph{Asymmetric Numeral
Systems} (ANS; \citealp{duda2009}), and also uses an idea called \emph{bits back
coding} \citep{wallace1990,hinton1993}. \citet{townsend2019} show that BB-ANS
can be used to compress sequences of symbols which are modeled as statistically
independent. The compression rate achieved for the first symbol compressed is
equal to the ELBO plus the entropy of the approximate posterior, which is a
significant overhead, but compression rates for subsequent symbols are very
close to the ELBO\@. For long enough sequences, the effect of the first symbol
overhead on the overall compression rate is negligible.

\citet{townsend2019} demonstrated BB-ANS on MNIST images using a small VAE.
Subsequent work has extended BB-ANS to hierarchical latent variable models and
larger, color images \citep{townsend2020,kingma2019} and demonstrated how to
use ideas from the Monte Carlo literature to achieve a tighter bound than the
ELBO \citep{ruan2021}.  In this work we present a generalization of BB-ANS to
sequences which are not modeled as independent, by `interleaving' bits-back
steps with the time-steps in a model, exploiting latent Markov structure in a
similar way to the Bit-Swap method of \citet{kingma2019}. We call the method
`\underline{I}nterleaving for e\underline{cono}mical \underline{C}ompression
with \underline{La}tent \underline{S}equence \underline{M}odels'
(IconoCLaSM)\@.

IconoCLaSM is applicable to the general class of `state space models' (SSMs;
defined in \cref{sec:ssms}), but does require access to specific conditional
distributions under an approximate posterior (\cref{sec:iconoclasm-method})\@.
We demonstrate the method on a hidden Markov model (HMM; \citealp{baum1966}),
showing that, as with vanilla BB-ANS, for long enough sequences the compression
rate is very close to the model ELBO\@. We believe that this result is
significant, because the same method is likely to be extensible to audio and
video compression with deep latent variable models.

\section{Background}
\subsection{Asymmetric Numeral Systems}
\label{sec:ans}
Asymmetric Numeral Systems (ANS) is a family of algorithms for losslessly
compressing sequences \citep{duda2009}. ANS defines a last-in-first-out (LIFO)
compressed message data structure and the inverse pair of functions `push' and
`pop', for adding and removing data from the message.

To encode a sequence \(x_1,\ldots,x_T\), with distribution \(P\), ANS encoding
starts with a very short base message \(m_\mathrm{init}\), and then encodes
the elements one-at-a-time, starting with \(x_T\) and working \emph{backwards}:
\begin{align}
m_T     &= \mathrm{push}_{P(x_T\given x_1,\ldots,x_{T-1})}(m_\mathrm{init}, x_T)\\
m_{T-1} &= \mathrm{push}_{P(x_{T-1}\given x_1,\ldots,x_{T-2})}(m_T, x_{T-1})\\
        &\kern5.5pt\vdots\nonumber\\
m_1     &= \mathrm{push}_{P(x_1)}(m_2, x_1).
\end{align}
The compressed message \(m_1\) can then be communicated and data decoded
forwards using the inverse sequence of pop operations. Each push depends on the
cumulative distribution function (CDF) and its inverse for the conditional
distribution \(P(x_t\given x_1,\ldots,x_{t-1})\).

It can be shown that the length, in bits, of the message after each push
operation is bounded above
\begin{equation}
  l(m_t) < h(x_t,\ldots,x_T\given x_1,\ldots,x_{t-1}) + t\epsilon + C,
\end{equation}
where \(h(x) := 1/\log_2 P(x)\) is the `information content', and \(C\) and
\(\epsilon\) are implementation-dependent constants.  Since \(C\) does not
depend on \(t\), the per-symbol compression rate tends towards \(h(x_t\given
x_1,\ldots,x_{t-1})+\epsilon\) as \(T\) increases. In a typical ANS
implementation, \(C = 64\) and \(\epsilon \approx 2.2\times 10^{-5}\);
\citet{townsend2020a} provides more detail.  Thus, for long enough sequences,
ANS encodes close to the information content, and it is not possible to do
better on average for data sampled from \(P\) \citep{shannon1948}.

After pushing some data, it is then possible to pop using a \emph{different}
distribution. Doing so produces a sample from the distribution used for
popping, and shortens the message length by the information content of that
sample. Using such intermediate decoding steps within an encoding process is
the key idea behind bits back coding, which is significantly more flexible than
basic ANS\@.

Hereafter we elide \(m\) and use the shorthand notation of
\citet{townsend2020}: \(x\rightarrow P(x)\) for pushing the symbol \(x\)
according to the distribution \(P\), and \(x\leftarrow P(x)\) for the inverse
pop operation.

\subsection{Latent variable models and state space models}\label{sec:ssms}
For the purposes of this work, we define a latent variable model to be a model
with a mass function which can be expressed as a sum or integral over a
`latent' variable
\(z\):
\begin{equation}
  P(x\given \theta) = \int P(x\given z, \theta)\, P(z\given \theta)\; dz,
\end{equation}
where \(\theta\) denotes the parameters of the model.  The marginal
distribution \(P(x\given\theta)\) needn't be tractable: our model is defined by
the `prior' \(P(z\given\theta)\) and `likelihood' \(P(x\given z, \theta)\).
The parameters \(\theta\) can be learned by optimizing the `evidence lower
bound' (ELBO):
\begin{equation}
  L(x; \theta, \phi) = \mathbb{E}_{Q(z\given x,\phi)}\log\frac{P(x,
  z\given\theta)}{Q(z\given x, \phi)},
\end{equation}
where \(Q\) is a variational distribution, referred to as the `approximate
posterior', and \(\phi\) are the `variational parameters'. The ELBO is
usually optimized jointly with respect to \(\theta\) and \(\phi\) using
stochastic gradient ascent \citep{kingma2014,rezende2014}.  Coding with BB-ANS
(\cref{sec:bb-ans}), requires access to conditional CDFs and inverse CDFs under
the prior, likelihood and approximate posterior, as described in \cref{sec:ans}.
From now on we assume access to a \emph{trained} model (i.e., with \(\theta\)
and \(\phi\) fixed).

State space models (SSMs) are latent variable models where an observed sequence
\(x_1,\ldots,x_T\) is modeled using a latent sequence \(z_1,\ldots,z_T\) and
the joint mass function admits the factorization
\begin{equation}
  P(x_1,\ldots,x_T, z_1,\ldots,z_T) = P(z_1)\prod_{t=2}^T P(z_t\given z_{t-1})
    \prod_{t=1}^T P(x_t\given z_t).
\end{equation}

In general, both the observations and the
latents may be either continuous or discrete, but for lossless compression we
require discrete observations. SSMs have been applied to speech modeling, computational
neuroscience and many other areas \citep{kalman1960,rabiner1989,paninski2010},
and using deep, nonlinear SSMs for video modeling is an active area of research
\citep{johnson2016,saxena2021}.

\subsection{Bits back with ANS}\label{sec:bb-ans}
To compress a symbol \(x\) using ANS with a latent variable model, a naive
thing to do would be to choose \(z\) and send it along with the data:
\begin{align}
  x&\rightarrow P(x\given z)\\
  z&\rightarrow P(z).
\end{align}
These operations increase message length by \(\log 1/P(x, z)\), so the optimal
choice of \(z\) is one which minimizes that quantity. That increase is longer
than the ideal length \(\log 1/P(x)\), because of a redundancy: \(z\) has
effectively been sent twice: once the receiver has decoded \(x\), they can
compute \(z\) by running the same minimization routine that the sender used.

The idea of bits-back coding \citep{wallace1990,hinton1993} is to recover
information communicated in the choice of \(z\), in a way that cancels out the
redundancy. BB-ANS \citep{townsend2019} is a practical realization of this
idea that codes at an average rate equal to the negative ELBO\@. The BB-ANS
sender \emph{decodes} \(z\) according to \(Q\).  That is, they sample a
plausible latent value \(z\) from the approximate posterior \(Q\), using the
information in the message to make the random choice.

The BB-ANS encoding and decoding processes are as follows:

\hspace{.05\textwidth}\begin{minipage}{.35\textwidth}
  \textbf{BB-ANS encoding process}
  \begin{lstlisting}
$z\leftarrow  Q(z\given x)$
$x\rightarrow P(x\given z)$
$z\rightarrow P(z)        $
  \end{lstlisting}
\end{minipage}
\hspace{.2\textwidth}
\begin{minipage}{.35\textwidth}
  \textbf{BB-ANS decoding process}
  \begin{lstlisting}
$z\leftarrow  P(z)        $
$x\leftarrow  P(x\given z)$
$z\rightarrow Q(z\given x)$
  \end{lstlisting}
\end{minipage}\hspace{.05\textwidth}
When the first symbol \(x\) is encoded, the message \(m\) is empty, so the
posterior sample~\(z\) is generated at random, rather than decoded.
This leads to an `initial bits' overhead of \(\log1/Q(z\given x)\) for the
first symbol. Once a sufficient buffer has been built up in \(m\), \(z\) can
be decoded and the compression rate of subsequent steps will be close to the
negative ELBO. If \(Q\) is equal to the exact posterior \(P(z\given x)\), then
the ELBO bound is tight and BB-ANS is close to optimal \citep{townsend2019}.

\section{Coding with state space models using IconoCLaSM}\label{sec:iconoclasm-method}
In order to use an SSM for lossless compression, we could directly apply
BB-ANS\@. However, the first step of encoding in BB-ANS is to decode an
entire latent, and in the case of an SSM that means sampling the sequence
\(z_1,\ldots,z_T\); leading to an initial bits overhead which scales
with \(T\), making this an impractical method. We could break the sequence into
independent chunks, but this would also harm the compression rate. The central
contribution of this work is a method for `interleaving' bits back steps with
the SSM timesteps, allowing optimal compression with \(O(1)\) initial bits
overhead. The encoding and decoding processes are shown below.

IconoCLaSM requires the following factorization of the approximate posterior:
\begin{equation}\label{eq:ssm-post-fac}
  Q(z_1,\ldots,z_T\given x_1,\ldots,x_T) = Q(z_T\given x_1,\ldots,x_T)
    \prod_{t=1}^{T-1}Q(z_t\given x_1,\ldots,x_t,z_{t+1}),
\end{equation}
and that the factors on the right hand side can be coded with ANS (i.e., their
CDFs and inverse CDFs are available). In HMMs and linear Gaussian SSMs those
factors can be computed exactly using message passing. In deep latent variable
models they may be computed using an RNN, as in \citet{saxena2021}, or by
message passing, as in \citet{johnson2016}.

\hspace{.025\textwidth}\begin{minipage}{.4\textwidth}
  \textbf{IconoCLaSM encoding process}
  \begin{lstlisting}
$z_T\leftarrow  Q(z_T\given x_1,\ldots,x_T)$
for $t$ in $T,\ldots,2$:
  $x_t\rightarrow P(x_t\given z_t)$
  $z_{t-1}\leftarrow Q(z_{t-1}\given x_1,\ldots,x_{t-1}, z_t)$
  $z_t\rightarrow P(z_t\given z_{t-1})$
$x_1\rightarrow P(x_1\given z_1)$
$z_1\rightarrow P(z_1)$
  \end{lstlisting}
\end{minipage}
\hspace{.1\textwidth}
\begin{minipage}{.4\textwidth}
  \textbf{IconoCLaSM decoding process}
  \begin{lstlisting}
$z_1\leftarrow P(z_1)$
$x_1\leftarrow P(x_1\given z_1)$
for $t$ in $2,\ldots,T$:
  $z_t\leftarrow P(z_t\given z_{t-1})$
  $z_{t-1}\rightarrow Q(z_{t-1}\given x_1,\ldots,x_{t-1}, z_t)$
  $x_t\leftarrow P(x_t\given z_t)$
$z_T\rightarrow  Q(z_T\given x_1,\ldots,x_T)$
  \end{lstlisting}
\end{minipage}\hspace{.025\textwidth}

\section{Experiments}
We conducted proof of concept experiments to demonstrate IconoCLaSM on a small
hidden Markov model; that is, an SSM with discrete latents and observations.
Since, in an HMM, the conditionals \(P(x_t\given x_1,\ldots,x_{t-1})\)
\emph{can} be tractably computed using message passing, it is possible to use
vanilla ANS with this model. As a result, IconoCLaSM is not the most
efficient way to code with an HMM. However, an HMM is a
reasonable first demonstration of IconoCLaSM, since the HMM
has the correct generative structure, the required posterior conditionals are
available, and we can easily compare to the exact information content.

\begin{figure}
  \centerline{\includegraphics{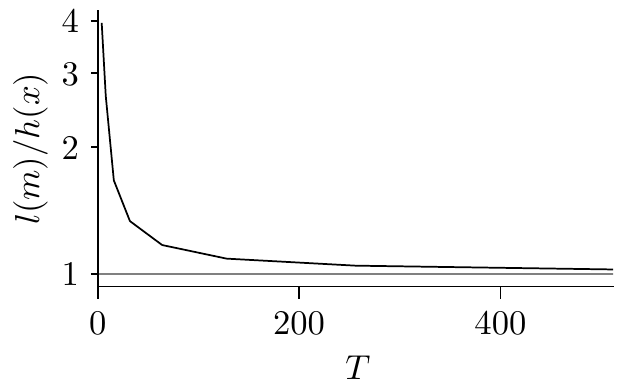} ~~ \includegraphics{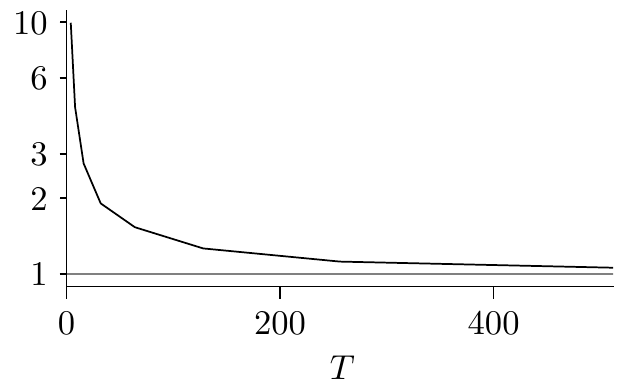}}
  \vspace*{-0.2cm}
  \caption{Plots showing that the ratio of the actual compressed message
    length, \(l(m)\), to the information content \(h(x)\), tends to \(1\) as
    the sequence length \(T\) increases. Left hand plot is compressing data
    sampled from the model; right hand plot is compressing English text with a
  model trained using the EM algorithm.}\label{fig:plots}
\end{figure}
\Cref{fig:plots} shows the convergence of the compression rate to the optimum,
\(h(x)\), as the sequence length \(T\) increases, for two different setups. For
the first experiment, shown in the left hand plot, we compressed an exact
sample from an HMM with randomly generated parameters. Then, to demonstrate
that IconoCLaSM doesn't depend on the model being perfectly fit to the data,
for the second plot we compressed a section of War and Peace, using an HMM
trained on an earlier section of the same text. The effect of the initial bits
overhead is significant for short messages, but diminishes gracefully as
sequence length increases.

The HMMs used for both experiments had 64 hidden states. The HMM for the first
experiment had 64 observation states and all parameters sampled from a
Dirichlet distribution with concentration parameters \(\alpha_i \!=\! 1\).
The HMM for the second experiment had 101 observation states (corresponding to
the characters that appeared in the text) and parameters trained using the EM
algorithm for 100 iterations. Code to reproduce these experiments is available
at \jurl{github.com/j-towns/ssm-code}.

\section{Discussion}
We have demonstrated the existence of a simple, practical scheme for doing
lossless compression with state space models, assuming the availability of
certain conditionals under a (possibly approximate) posterior. It was
demonstrated by \citet{kingma2019} and \citet{townsend2020} that the basic
BB-ANS method can be scaled up to large, color images. We speculate that it may
be possible to scale up IconoCLaSM in a similar way, and use it for lossless
compression of video, where deep latent variable models have been shown to be
effective \citep{saxena2021,johnson2016}. Two particularly interesting open
questions are
\begin{enumerate}
  \item Can IconoCLaSM be generalized to hierarchical models, and if so what
    constraints are there on the hierarchical model and posterior?
  \item Can IconoCLaSM be combined with Monte Carlo methods such as sequential
    Monte Carlo in order to improve on the ELBO, in the style of
    \citet{ruan2021}?
\end{enumerate}
We leave investigation of these questions to future work.

\bibliography{iclr2021_conference}
\bibliographystyle{iclr2021_conference}

\end{document}